\documentclass[10pt,logo,copyright]{giga-report}
\usepackage[authoryear,sort&compress,round]{natbib}

\usepackage[utf8]{inputenc} %
\usepackage[T1]{fontenc}    %

\usepackage{parskip}        %
\usepackage{url}            %
\usepackage{booktabs}       %
\usepackage{amsfonts}       %
\usepackage{nicefrac}       %
\usepackage{microtype}      %
\usepackage{xcolor}         %
\usepackage[dvipsnames]{xcolor} %
\usepackage{graphicx}
\usepackage{animate}        %
\usepackage{subcaption}
\usepackage{tabularx}
\usepackage{makecell}
\usepackage{adjustbox}
\usepackage{setspace}
\usepackage{todonotes}
\usepackage{colortbl}
\usepackage{wrapfig}

\newcolumntype{M}[1]{>{\centering\arraybackslash}m{#1}}
\usepackage{float}
\usepackage{tikz}
\usetikzlibrary{positioning,shapes,arrows}
\usepackage{amsmath,amsfonts,bm, bbm,leftindex}
\usepackage{multirow}
\usepackage{comment}
\usepackage{gensymb}
\usepackage{lipsum}
\usetikzlibrary{arrows.meta, positioning, fit}
\usepackage[para]{threeparttable}
\usepackage{tikz}
\usetikzlibrary{tikzmark}

\def\eqref#1{equation~\ref{#1}}

\def\1{\bm{1}}

\DeclareMathAlphabet{\mathsfit}{\encodingdefault}{\sfdefault}{m}{sl}
\SetMathAlphabet{\mathsfit}{bold}{\encodingdefault}{\sfdefault}{bx}{n}

\makeatletter
\let\save@mathaccent\mathaccent
\newcommand*\if@single[3]{%
  \setbox0\hbox{${\mathaccent"0362{#1}}^H$}%
  \setbox2\hbox{${\mathaccent"0362{\kern0pt#1}}^H$}%
  \ifdim\ht0=\ht2 #3\else #2\fi
  }
\newcommand*\rel@kern[1]{\kern#1\dimexpr\macc@kerna}
\newcommand*\widebar[1]{\@ifnextchar^{{\wide@bar{#1}{0}}}{\wide@bar{#1}{1}}}
\newcommand*\wide@bar[2]{\if@single{#1}{\wide@bar@{#1}{#2}{1}}{\wide@bar@{#1}{#2}{2}}}
\newcommand*\wide@bar@[3]{%
  \begingroup
  \def\mathaccent##1##2{%
    \let\mathaccent\save@mathaccent
    \if#32 \let\macc@nucleus\first@char \fi
    \setbox\z@\hbox{$\macc@style{\macc@nucleus}_{}$}%
    \setbox\tw@\hbox{$\macc@style{\macc@nucleus}{}_{}$}%
    \dimen@\wd\tw@
    \advance\dimen@-\wd\z@
    \divide\dimen@ 3
    \@tempdima\wd\tw@
    \advance\@tempdima-\scriptspace
    \divide\@tempdima 10
    \advance\dimen@-\@tempdima
    \ifdim\dimen@>\z@ \dimen@0pt\fi
    \rel@kern{0.6}\kern-\dimen@
    \if#31
      \overline{\rel@kern{-0.6}\kern\dimen@\macc@nucleus\rel@kern{0.4}\kern\dimen@}%
      \advance\dimen@0.4\dimexpr\macc@kerna
      \let\final@kern#2%
      \ifdim\dimen@<\z@ \let\final@kern1\fi
      \if\final@kern1 \kern-\dimen@\fi
    \else
      \overline{\rel@kern{-0.6}\kern\dimen@#1}%
    \fi
  }%
  \macc@depth\@ne
  \let\math@bgroup\@empty \let\math@egroup\macc@set@skewchar
  \mathsurround\z@ \frozen@everymath{\mathgroup\macc@group\relax}%
  \macc@set@skewchar\relax
  \let\mathaccentV\macc@nested@a
  \if#31
    \macc@nested@a\relax111{#1}%
  \else
    \def\gobble@till@marker##1\endmarker{}%
    \futurelet\first@char\gobble@till@marker#1\endmarker
    \ifcat\noexpand\first@char A\else
      \def\first@char{}%
    \fi
    \macc@nested@a\relax111{\first@char}%
  \fi
  \endgroup
}
\makeatother

\definecolor{darkred}{rgb}{0.7, 0.0, 0.0}

\usepackage{amsmath} 

\usepackage[table]{xcolor}

\usepackage{pifont}
\usepackage{graphicx}

\usepackage{marvosym}
\usepackage[nameinlink]{cleveref}
\crefname{equation}{Eq.}{Eqs.}
\crefname{figure}{Fig.}{Figs.}
\crefname{section}{Sec.}{Sec.}
\crefname{appendix}{App.}{App.}
\crefname{table}{Tab.}{Tabs.}
\crefname{algorithm}{Algo}{Algo}
\crefname{thm}{Thm}{Thm}
\Crefname{thm}{Thm}{Thm}
\crefname{prop}{Prop}{Prop}

\newcommand{\crefnames}[3]{%
  \@for\next:=#1\do{%
    \expandafter\crefname\expandafter{\next}{#2}{#3}%
  }%
}

\title{\centering ReconPhys: Reconstruct Appearance and Physical Attributes from Single Video}
\author{
\vspace{-0.1in}

\footnotesize

\normalfont
\centerline{
Boyuan Wang\textsuperscript{\rm 1,\rm 2},
Xiaofeng Wang\textsuperscript{\rm 1,\rm 3},
Yongkang Li\textsuperscript{\rm 1},
Zheng Zhu\textsuperscript{\rm 1 \Letter},
Yifan Chang\textsuperscript{\rm 2},
Angen Ye\textsuperscript{\rm 2}

}

\centerline{
Guosheng Zhao\textsuperscript{\rm 1,\rm 2},
Chaojun Ni\textsuperscript{\rm 1},
Guan Huang\textsuperscript{\rm 1},
Yijie Ren\textsuperscript{\rm 2},
Yueqi Duan\textsuperscript{\rm 3},
Xingang Wang\textsuperscript{\rm 2 \Letter}
}

\centerline{\textsuperscript{\rm 1}GigaAI~~~~\textsuperscript{\rm 2}Institute of Automation, Chinese Academy of Sciences~~~~\textsuperscript{\rm 3}Tsinghua University} 

\centerline{{Project Page: \href{https://chuanshuogushi.github.io/ReconPhys/}{https://chuanshuogushi.github.io/ReconPhys}}} 

\vspace{-1em}
}

\begin{document}
\maketitle

\begin{figure*}[htbp]
    \centering
    \resizebox{\textwidth}{!}{
        \includegraphics{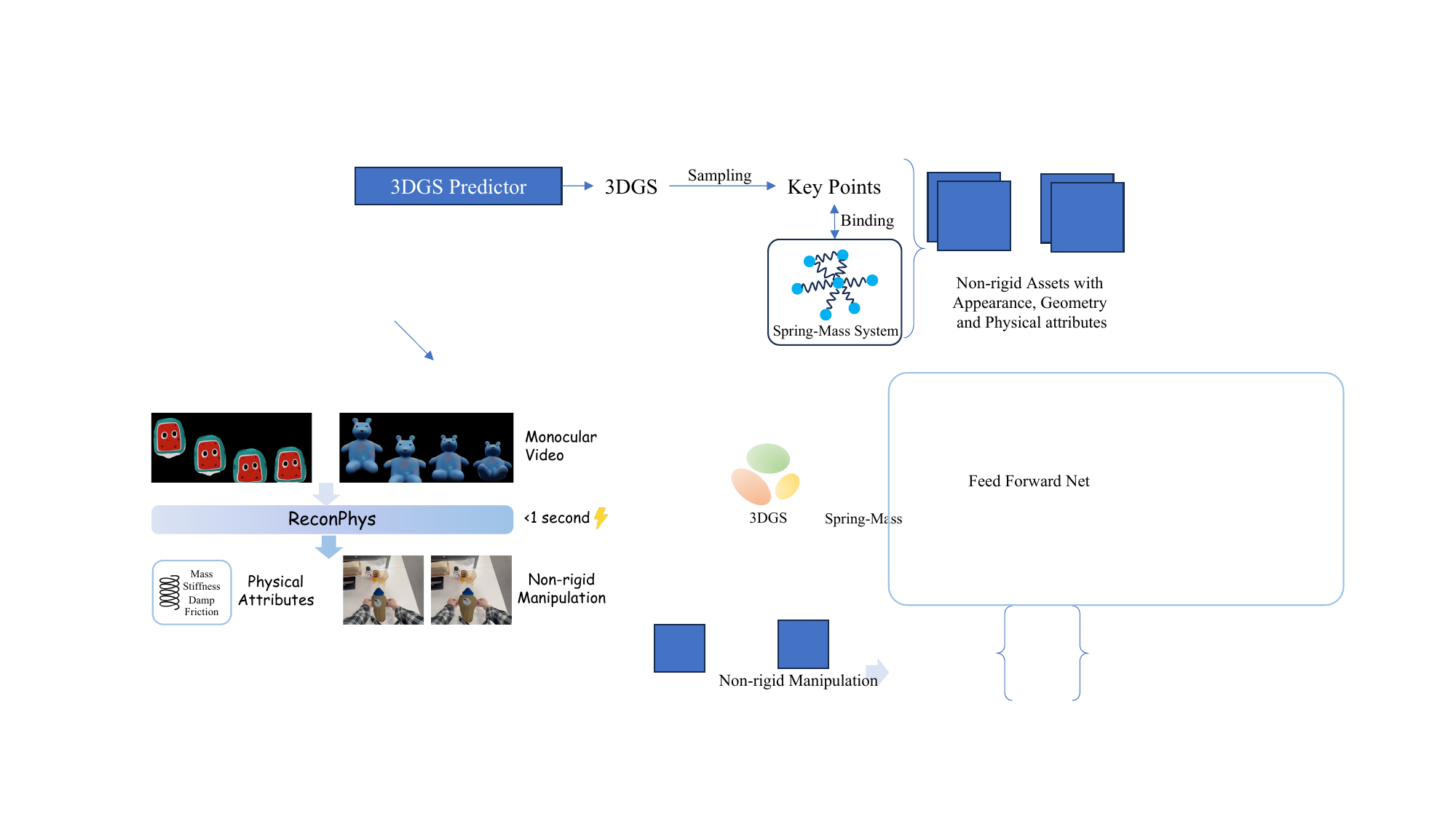}
    }
    \caption{ReconPhys predicts physical attributes using a feedforward neural network (FFN). This approach obviates the need for per-scene optimization and requires only a monocular video of the object as input. Within one second, it produces 3DGS assets with binded physical attributes. This capability offers a promising and efficient pathway for generating simulation-ready data for non-rigid manipulation, enabling the rapid transfer of real-world objects into simulation environments without human intervention.}
    \label{fig:main_demo}
\end{figure*}

\begin{abstract}
\vspace{-0.1in}
Reconstructing non-rigid objects with physical plausibility remains a significant challenge. Existing approaches leverage differentiable rendering for per-scene optimization, recovering geometry and dynamics but requiring expensive tuning or manual annotation, which limits practicality and generalizability. To address this, we propose ReconPhys, the first feedforward framework that jointly learns physical attribute estimation and 3D Gaussian Splatting reconstruction from a single monocular video. Our method employs a dual-branch architecture trained via a self-supervised strategy, eliminating the need for ground-truth physics labels. Given a video sequence, ReconPhys simultaneously infers geometry, appearance, and physical attributes. Experiments on a large-scale synthetic dataset demonstrate superior performance: our method achieves 21.64 PSNR in future prediction compared to 13.27 by state-of-the-art optimization baselines, while reducing Chamfer Distance from 0.349 to 0.004. Crucially, ReconPhys enables fast inference ($<$1 second) versus hours required by existing methods, facilitating rapid generation of simulation-ready assets for robotics and graphics.
\end{abstract}

\abscontent
\section{Introduction}
Non-rigid object reconstruction lies at the intersection of computer vision, graphics, and physics-based simulation, with applications ranging from virtual reality and robotics to digital content creation \cite{neuralgraphs,npgs,pacnerf,tretschk2023state,4dgs,dygs,gs-dynamics,springgaus,phystwin,omniphysgs}. Unlike rigid bodies, non-rigid objects undergo complex deformations under external forces or internal dynamics, making their 3D recovery from 2D observations a highly ill-posed and challenging inverse problem. Accurately capturing not only their geometry and appearance but also their underlying physical behavior is crucial for enabling realistic interaction and simulation in downstream tasks. As illustrated in Figure~\ref{fig:main_demo}, our framework directly addresses this need: by processing a single monocular video, it generates simulation-ready 3DGS assets with bound physical attributes within one second, eliminating the need for per-scene optimization.

Recent advances in dynamic scene reconstruction have been driven by Neural Radiance Fields (NeRF) \cite{nerf} and, more recently, 3D Gaussian Splatting (3DGS) \cite{3dgs}. Early dynamic NeRF variants, such as D-NeRF \cite{d-nerf} and Nerfies \cite{nerfies}, model temporal changes via canonical-space deformation or scene flow, achieving impressive view synthesis quality but often incurring high training and rendering costs that limit interactive usage. The emergence of 3DGS has revolutionized this field by offering explicit representations with real-time differentiable rendering. Dynamic 3DGS methods, including deformation-based approaches \cite{deformable-gs,dygs} and explicit per-timestep modeling \cite{4dgs,gaussianflow}, can render photorealistic dynamic videos at high speed. However, despite their visual fidelity, these methods primarily optimize geometry and appearance without explicitly recovering physically meaningful parameters. Consequently, their predictions may be visually plausible yet physically inconsistent under unseen forces or interactions, motivating the integration of physics-aware modeling into dynamic reconstruction.

Despite these advances, current techniques suffer from critical limitations that hinder scalability and generalization. Bridging visual reconstruction with physical understanding is crucial for simulation-ready asset creation. Representative directions couple neural representations with differentiable physics engines, such as PAC-NeRF \cite{pacnerf} which integrates Material Point Method (MPM) dynamics but assumes known material families, limiting adaptability to heterogeneous objects. Similarly, PhysGaussian \cite{physgaussian} incorporates continuum mechanics into 3DGS but focuses on generative dynamics rather than inferring per-scene attributes from observations. The closest work, Spring-Gaus \cite{springgaus}, binds a spring-mass system to 3D Gaussians for elastic objects. However, it relies on dense multi-view inputs and expensive per-scene optimization, making it impractical for general monocular deployment. This reliance on per-scene optimization not only limits practical deployment but also fails to leverage shared physical priors across object categories or materials.

To address these limitations, we propose ReconPhys, which integrates 3DGS with a differentiable spring-mass system for end-to-end inference. Unlike prior approaches that decouple visual reconstruction from physical modeling, our framework enables joint estimation of geometry, appearance, and material properties without manual annotation or per-scene optimization.

Our core insight is to replace scene-specific physics tuning with a feedforward neural estimator trained on diverse dynamic observations. By conditioning physical attributes prediction on reconstructed shape and observed motion dynamics, ReconPhys learns generalizable mappings from visual cues to material properties, allowing zero-shot transfer to unseen objects. The entire pipeline is trained self-supervised via photometric loss, leveraging differentiable rendering and physics simulation to propagate gradients from image-space errors back to physical attributes.

Extensive experiments demonstrate that ReconPhys achieves state-of-the-art performance in both dynamic reconstruction and future-state prediction. Our method not only improves visual fidelity (PSNR, LPIPS) but also yields a dramatic improvement in 3D geometric accuracy, reducing Chamfer Distance (CD) from 0.593 to 0.001 compared to 4DGS \cite{4dgs}. Notably, in future prediction, our approach vastly outperforms Spring-Gaus \cite{springgaus} across all metrics (e.g., 21.64 vs. 13.27 PSNR), confirming that embedding physical understanding directly into the reconstruction process not only enhances visual fidelity but also unlocks practical applications in robotics and simulation where realistic interaction is essential.

Our key contributions are threefold:

(1) We present the first feedforward, general-purpose, and fully differentiable framework that jointly reconstructs non-rigid object geometry, appearance, and physical attributes from a single video, enabling rapid inference without per-scene optimization;

(2) We design a novel dual-branch architecture with a self-supervised physics training strategy and contribute an automated pipeline to synthesize a large-scale dataset of deformable objects with diverse physical attributes;

(3) We demonstrate through extensive experiments that our method achieves state-of-the-art performance in both dynamic reconstruction and future-state prediction. Specifically, we yield a dramatic improvement in 3D geometric accuracy (reducing CD from 0.5932 to 0.001 compared to 4DGS) and vastly outperform Spring-Gaus in future prediction, enabling physically plausible simulation of unseen deformations.
\section{Related Works}

\subsection{Dynamic 3DGS Reconstruction}
Dynamic scene reconstruction has rapidly advanced with neural rendering, most notably Neural Radiance Fields (NeRF)~\cite{nerf}. Beyond the canonical static setting, dynamic NeRF variants model time-varying scenes via canonical-space deformation or scene flow, enabling monocular dynamic view synthesis~\cite{d-nerf,nerfies,hypernerf,nsff,nonrigidnerf,flowsup_deformnerf,driess2023learning}. Despite strong visual quality, NeRF-based dynamics are often computationally expensive and can be slow at render time, which limits interactive usage.

Recently, 3D Gaussian Splatting (3DGS)~\cite{3dgs} has emerged as an explicit representation that achieves real-time differentiable rendering, inspiring a line of Dynamic 3DGS methods. Existing approaches can be broadly grouped into deformation-based and explicit modeling strategies. Deformation-based methods learn a canonical set of Gaussians and a deformation field to animate the scene, improving stability and temporal coherence~\cite{deformable-gs,dygs}. In contrast, explicit approaches directly optimize Gaussian positions and attributes over time, or introduce spatio-temporal parameterizations to better capture non-rigid motion~\cite{4dgs,gaussianflow,dyntgs,dynmf,scgs,cogs,humandreamer-x,jiang2024vr,guo2024motion,zheng2023avatarrex}. While these methods can render photorealistic dynamic videos at high speed, they primarily target geometric and appearance fidelity and do not explicitly recover physically meaningful parameters. As a result, their predictions may be visually plausible yet physically inconsistent under unseen forces or interactions, motivating physics-aware modeling for simulation-ready dynamic assets.

\subsection{Physics-Aware Reconstruction and Physical Attributes Prediction}
Bridging visual reconstruction with physical understanding is crucial for simulation-ready asset creation. A representative direction couples neural representations with differentiable physics engines, enabling gradients from image-space losses to flow back to physical parameters. PAC-NeRF~\cite{pacnerf} integrates differentiable Material Point Method (MPM) dynamics into a NeRF-like representation for geometry-agnostic system identification, but assumes known material families and typically uses coarse global parameters, limiting adaptability to heterogeneous real objects. PhysGaussian~\cite{physgaussian} incorporates continuum mechanics into 3DGS to generate physically grounded motion, yet it is primarily designed for generative dynamics rather than directly inferring per-scene physical attributes from monocular observations.

Another line of work focuses on learning-based physical modeling~\cite{pinn,gnsimulator,nclaw,diffpd,simplerenv,embodiedreamer}. These methods provide powerful tools for modeling dynamics, but often require either explicit state supervision, expensive per-scene optimization, or specialized simulation data.

The closest to our setting is Spring-Gaus~\cite{springgaus}, which binds a spring-mass system to 3D Gaussians for reconstructing and simulating elastic objects. However, it requires multi-view capture and relies on per-scene optimization, making it impractical for general monocular deployment and large-scale usage. In contrast, our approach targets feedforward and monocular physical attribute inference: we combine 3DGS reconstruction with a differentiable dynamics model, and train a neural estimator to predict physical parameters directly from video, removing the need for per-scene optimization or manual physics annotation, enabling zero-shot generalization to unseen objects.

\section{Method}
\subsection{Problem Formulation}
\label{sec:problem_formulation}
Given a monocular video $\mathcal{V}$ capturing an object $\mathcal{O}$ undergoing gravitational drop, collision, and rebound dynamics, we aim to recover its intrinsic physical attributes and 3D visual representation solely from observed motion.
The object is modeled as a deformable body governed by a spring-mass system.
Let $\{\mathbf{x}_i\}_{i=1}^{N_A}$ denote the $N_A$ mass points, whose pairwise connectivity is determined once via $K$-nearest neighbors (KNN) on the initial shape and remains fixed thereafter.
The physical attributes are defined as $\mathbf{p} = (\{m_i\}, \{k_{ij}\}, \{d_{ij}\}, f)$, where $m_i$ is the mass of point $i$, $k_{ij} \geq 0$ and $d_{ij} \geq 0$ are the stiffness and damping coefficients of the spring connecting points $i$ and $j$, and $f \geq 0$ is the global friction coefficient between the object and the ground.

The geometry and appearance of the object are represented by a 3DGS representation comprising $N$ Gaussian kernels.
Crucially, every Gaussian kernel $\mathbf{g}_j$ is bound to the spring-mass system, such that its 3D center $\boldsymbol{\mu}_j$ is fully determined by the simulated mass-point states.
Each kernel is further characterized by orientation $\boldsymbol{\theta}_j$, RGB color $\mathbf{c}_j$, scale $\boldsymbol{\sigma}_j$, and opacity $\alpha_j$.

Our feedforward prediction model $\mathcal{M}$ takes $\mathcal{V}$ as input and jointly regresses both components:
\begin{equation}
    (\hat{\mathbf{p}}, \hat{\mathbf{g}}) = \mathcal{M}(\mathcal{V}),
\end{equation}
enabling holistic inference of appearance, geometry, and physics from monocular observations.
The initial mass point configuration $\{\mathbf{x}_i^{0}\}_{i=1}^{N_A}$ is derived from the first frame of $\mathcal{V}$, establishing the reference state for physical simulation.

\subsection{Physical Dynamics Model}
\label{sec:preliminary}

\textbf{3D Spring-Mass System.}
We model elastic object dynamics using a 3D spring-mass system composed of $N_A$ anchor points $\mathcal{A} = \{\mathbf{x}_i\}_{i=1}^{N_A}$, where each anchor $\mathbf{x}_i \in \mathbb{R}^3$ has mass $m_i$ and velocity $\mathbf{v}_i$.
Following Spring-Gaus~\cite{springgaus}, we establish spring connections between each anchor and its $K$ nearest neighbors through K-nearest neighbors (KNN) search:
\begin{equation}
    \mathcal{L} = \{l_{i,j}\}_{i=1,j=1}^{N_A,K} = \text{knn}(\mathcal{A}, \mathcal{A}, K),
\end{equation}
where $l_{i,j}$ denotes the rest length (initial distance) between anchors $\mathbf{x}_i$ and $\mathbf{x}_{i,j}$.
Each spring is characterized by stiffness $k_{i,j}$ and damping coefficient $d_{i,j}$.

The net force acting on anchor $\mathbf{x}_i^t$ at timestep $t$ combines spring forces, damping forces, and gravity:
\begin{equation}
    \mathbf{F}_i^t
    = \sum_{j=1}^{K} \mathbf{F}_{i,j}^{k,t}
    + \sum_{j=1}^{K} \mathbf{F}_{i,j}^{d,t}
    + m_i\mathbf{g}_{\mathrm{grav}},
    \label{eq:force}
\end{equation}
where the nonlinear spring force follows a generalized Hooke's law with exponent $p_k$:
\begin{equation}
    \mathbf{F}_{i,j}^{k,t}
    = -k_{i,j}\left(\|\mathbf{x}_i^t-\mathbf{x}_{i,j}^t\|-l_{i,j}\right)^{p_k}
    \cdot \frac{\mathbf{x}_i^t-\mathbf{x}_{i,j}^t}{\|\mathbf{x}_i^t-\mathbf{x}_{i,j}^t\|},
\end{equation}
and the damping force is:
\begin{equation}
    \mathbf{F}_{i,j}^{d,t}
    = -d_{i,j}\,(\mathbf{v}_i^t-\mathbf{v}_{i,j}^t)\cdot
    \frac{\mathbf{x}_i^t-\mathbf{x}_{i,j}^t}{\|\mathbf{x}_i^t-\mathbf{x}_{i,j}^t\|}
    \cdot
    \frac{\mathbf{x}_i^t-\mathbf{x}_{i,j}^t}{\|\mathbf{x}_i^t-\mathbf{x}_{i,j}^t\|}.
\end{equation}

\noindent \textbf{Numerical Integration.}
Anchor positions and velocities are updated via semi-implicit Euler integration with timestep $\Delta t$:

\begin{align}
    \hat{\mathbf{v}}_i^{t+1} &= \mathbf{v}_i^t + \frac{\mathbf{F}_i^t}{m_i} \Delta t, \\
    \hat{\mathbf{x}}_i^{t+1} &= \mathbf{x}_i^t + \hat{\mathbf{v}}_i^{t+1} \Delta t,
    \label{eq:integration}
\end{align}
followed by boundary condition enforcement $\mathcal{B}(\cdot)$ to model ground collisions:
\begin{equation}
    (\mathbf{x}_i^{t+1}, \mathbf{v}_i^{t+1}) = \mathcal{B}(\hat{\mathbf{x}}_i^{t+1}, \hat{\mathbf{v}}_i^{t+1}).
\end{equation}

\noindent \textbf{Binding to 3D Gaussian Splatting.}
To bridge physical simulation with visual representation, we introduce a two-stage binding mechanism between the spring-mass system and 3DGS:

1) Anchor Sampling:
Given $N$ Gaussian centers $\mathcal{X} = \{\boldsymbol{\mu}_i\}_{i=1}^N$, we perform volume sampling to generate $N_A$ anchor points ($N_A \ll N$ for efficiency):
\begin{equation}
    \mathcal{A} = \mathcal{V}_{\mathrm{vol}}(\mathcal{X}),
\end{equation}
where $\mathcal{V}_{\mathrm{vol}}(\cdot)$ is a sampling function that distributes anchors throughout the object volume rather than just on the surface, ensuring stable simulation.

2) Position Interpolation:
After simulating anchor dynamics to timestep $t+1$, Gaussian centers are updated via Inverse Distance Weighting (IDW) interpolation using their $n_b$ nearest anchors:
\begin{equation}
    \boldsymbol{\mu}_i^{t+1}
    = \frac{\sum_{j=1}^{n_b} \mathbf{x}_{i,j}^{t+1} \cdot (1/r_{i,j}^{p_b})}
    {\sum_{j=1}^{n_b} (1/r_{i,j}^{p_b})},
\end{equation}
where $r_{i,j} = \|\boldsymbol{\mu}_i^0 - \mathbf{x}_{i,j}^0\|$ is the initial distance between Gaussian center $\boldsymbol{\mu}_i$ and anchor $\mathbf{x}_{i,j}$, and $p_b > 0$ controls distance falloff.
This binding preserves local deformation patterns while enabling efficient simulation with sparse anchors.

\begin{figure*}[htbp]
    \centering
    \resizebox{\textwidth}{!}{
        \includegraphics{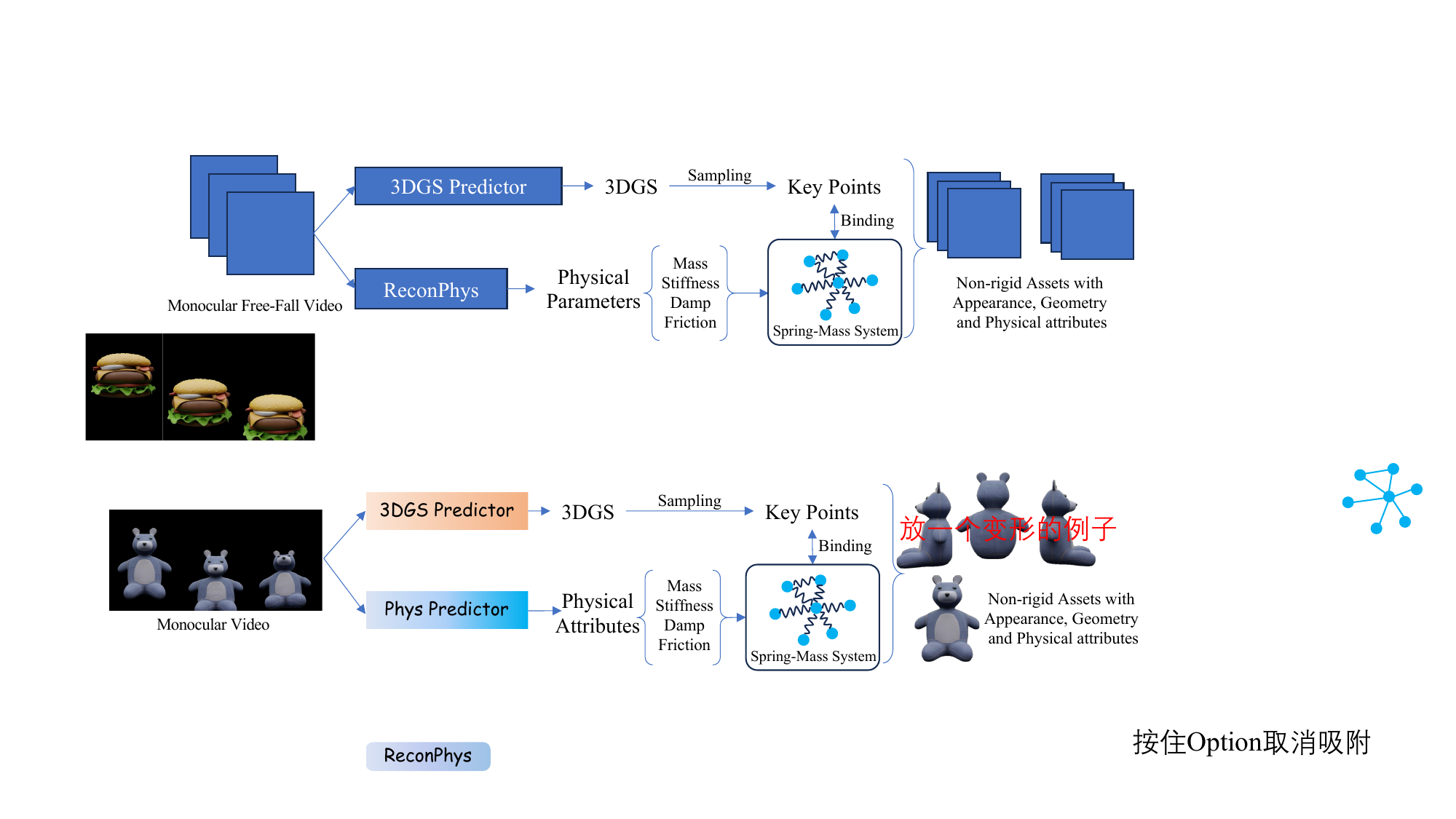}
    }
    \caption{Overview of our framework. Given an input image, a 3DGS predictor reconstructs the 3D object and samples anchor points to build a spring--mass system. Meanwhile, a physics predictor estimates physical attributes from videos to parameterize the system, producing simulatable 3D objects.}
    \label{fig:framework}
\end{figure*}

\subsection{Model Architecture}
\label{sec:arch}

As illustrated in Figure~\ref{fig:framework}, ReconPhys contains two components: a 3DGS predictor that provides a high-fidelity canonical representation for appearance and geometry, and a physical predictor that estimates compact physical attributes of a spring-mass system.
Given a monocular video $\mathcal{V}=\{I_t\}_{t=1}^{T}$, the 3DGS predictor outputs a canonical set of Gaussians $\hat{\mathbf{g}}^{0}$ (including centers, scale, rotation, color, and opacity), while the physical predictor outputs $\hat{\mathbf{p}}=(\hat m,\hat k,\hat d,\hat f)$ corresponding to mass, stiffness, damping, and friction.
For consistency with $\mathbf{p}=(\{m_i\},\{k_{ij}\},\{d_{ij}\},f)$, we use shared parameters across points/springs, i.e., $\hat m_i=\hat m$, $\hat k_{ij}=\hat k$, and $\hat d_{ij}=\hat d$.
Importantly, we directly use the off-the-shelf pretrained weights of the 3DGS predictor and keep it \emph{frozen} throughout training; thus, our learning focuses on identifying physical attributes and their simulation-consistent deformations.

The physical predictor employs an InternViT-based vision encoder to extract dynamic features from each frame of $\mathcal{V}$.
These per-frame features are subsequently processed by a ResNet \cite{resnet} backbone integrated with a self-attention mechanism \cite{vaswani2017attention} to aggregate spatio-temporal context, yielding a compact physical feature representation.
Finally, a dedicated MLP-based physical decoder maps these features to the estimated physical attributes $\hat{\mathbf{p}}$.
These attributes parameterize a differentiable spring-mass simulator that evolves anchor states over time.
The simulated anchor motion is then transferred to the Gaussian representation via the binding/interpolation mechanism in \cref{sec:preliminary}, yielding a time-varying 3DGS $\hat{\mathbf{g}}^{t}$ whose Gaussian centers $\boldsymbol{\mu}^{t}$ follow the simulated deformation.
This design establishes a tight coupling \emph{during training}: the video reconstruction error supervises physical attributes through a differentiable simulation--rendering loop.
At inference time, the predicted $\hat{\mathbf{p}}$ can be exported as a simulation-ready asset, while $\hat{\mathbf{g}}^{0}$ serves as a stable canonical visual representation.

\begin{figure*}[htbp]
    \centering
    \resizebox{0.9\textwidth}{!}{
        \includegraphics{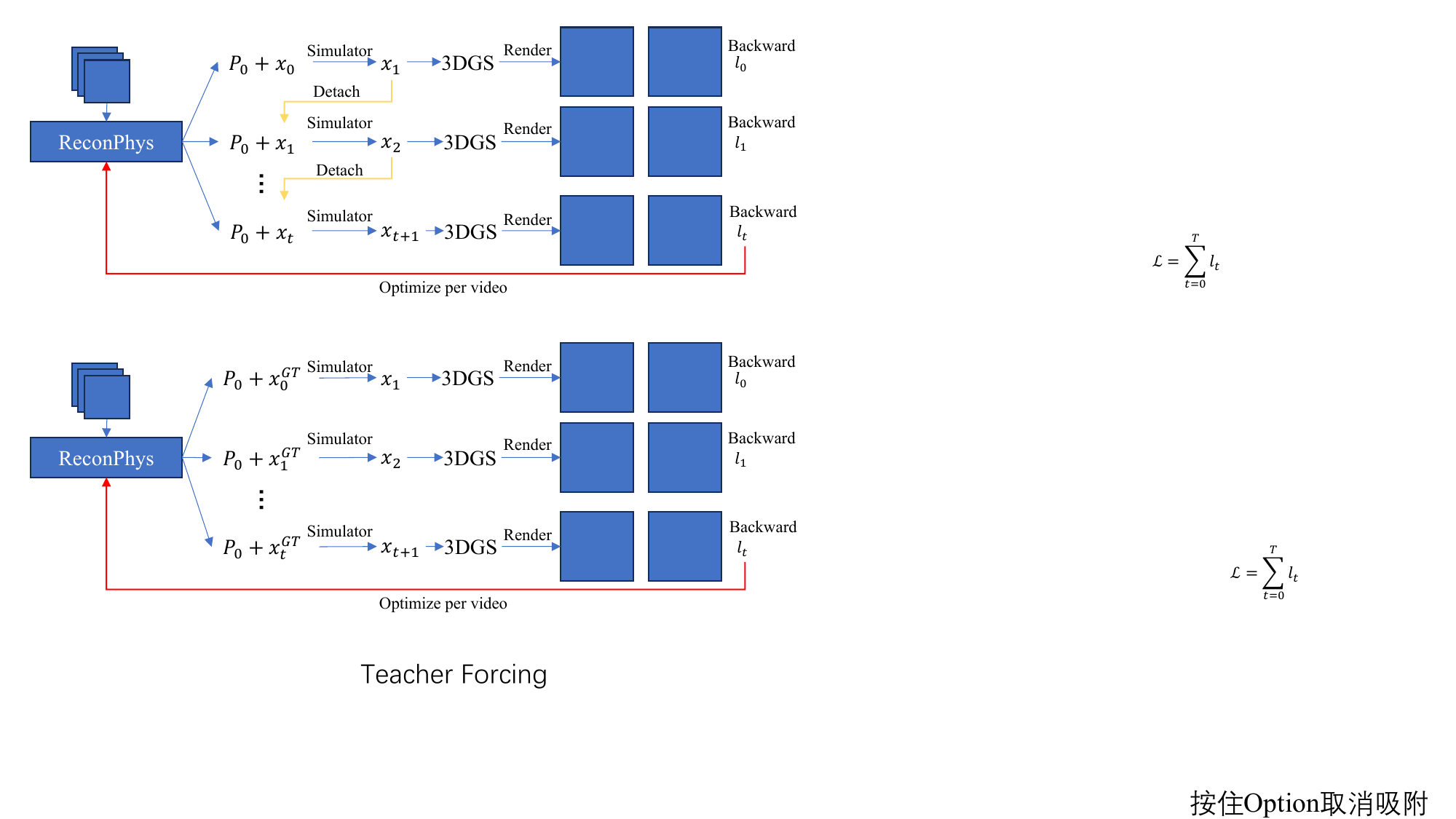}
    }
    \caption{Training pipeline of ReconPhys with self-forcing. The physics predictor estimates physical parameters from input videos to drive a spring–mass system for motion prediction. The predicted state is rendered and supervised by a reconstruction loss, with a self-forcing feedback loop for temporal prediction.}
    \label{fig:train_pipe}
\end{figure*}

\subsection{Differentiable Simulation--Rendering Loop}
\label{sec:diff_loop}

To learn $\hat{\mathbf{p}}$ without ground-truth physical annotations, we build an end-to-end differentiable computation graph that connects pixel-level reconstruction errors to physical attributes as shown in Figure~\ref{fig:train_pipe}.
In the forward pass, the simulator takes $\hat{\mathbf{p}}$ and recursively predicts anchor states $\{ \mathbf{x}_i^t, \mathbf{v}_i^t \}$ using differentiable force computation and numerical integration.
The anchor displacements are propagated to Gaussian centers through the binding rule, producing deformed centers $\boldsymbol{\mu}^{t}$ and hence a deformed 3DGS $\hat{\mathbf{g}}^{t}$.
A differentiable 3DGS renderer then generates the predicted frame $\hat I_t$ from $\hat{\mathbf{g}}^{t}$.

In the backward pass, the reconstruction loss provides gradients from $\hat I_t$ to $\boldsymbol{\mu}^{t}$ through the renderer.
Since $\boldsymbol{\mu}^{t}$ is deterministically induced by anchor states via binding, gradients further propagate to $\mathbf{x}_i^t$.
Because the simulator is differentiable, gradients traverse the unrolled dynamics and reach the physical parameters $\hat{\mathbf{p}}$, thereby updating the physical predictor.
Notably, since the 3DGS predictor is frozen, this gradient pathway is dedicated to identifying physically meaningful attributes that best explain the observed motion, rather than improving the underlying 3DGS reconstruction.

\subsection{Self-Supervised Training with Self-Forcing}
\label{sec:training}

We train ReconPhys in a self-supervised manner using only monocular videos.
Given $\mathcal{V}$, the model predicts $\hat{\mathbf{p}}=\mathcal{M}_p(\mathcal{V})$ and performs an autoregressive simulation rollout to obtain anchor trajectories, which deform the canonical Gaussians and produce rendered frames $\{\hat I_t\}$.
To reduce the training--inference mismatch, we adopt Self Forcing~\cite{selfforcing}: at each step, the simulator advances from the model's own previous predicted state instead of relying on recovered or proxy ground-truth states.
This encourages long-horizon stability and improves robustness to accumulated errors.

Autoregressive simulation may lead to unstable gradients when backpropagating through long unrolled trajectories.
We therefore apply truncated backpropagation by detaching the input state of each rollout step before computing $\mathbf{x}_{t+1}$.
This prevents gradient explosion or vanishing while still allowing supervision signals from rendered frames to update $\hat{\mathbf{p}}$.
During rollout, we update Gaussian centers according to the simulated deformation while keeping appearance-related attributes (rotation, color, scale, opacity) fixed, consistent with using a frozen canonical 3DGS.
The training objective is a photometric reconstruction loss:
\begin{equation}
\mathcal{L}=\sum_{t=1}^{T}\|I_t-\hat I_t\|_2^2,
\label{eq:recon_loss}
\end{equation}
which is fully differentiable with respect to $\hat{\mathbf{p}}$.
Minimizing Eq.~\eqref{eq:recon_loss} enables the model to identify physically plausible parameters that reproduce the observed deformation dynamics without any physical labels.

\section{Experiment}
\label{sec:experiments}

\subsection{Synthetic Data Pipeline}
\label{sec:data_synthesis}
Learning-based physical reconstruction often suffers from a scarcity of large-scale datasets pairing 3D assets with ground-truth physical attributes. To address this, we establish an automated pipeline to synthesize a dataset comprising 3DGS assets and their corresponding dynamics.

\noindent \textbf{Semantic Selection and Asset Generation.}
We source 3D assets from the Objaverse-XL collection~\cite{objaverse-xl}. To ensure suitability for non-rigid dynamics, we employ Qwen3-8B~\cite{qwen3} to filter candidates based on semantic labels, resulting in 500 eligible objects. For each object, we render four orthogonal views and utilize TRELLIS~\cite{trellis} to reconstruct high-fidelity 3DGS representations, serving as the initial geometric kernels.

\noindent \textbf{Consistent Anchor Sampling.}
To ensure the physical representation is reproducible and comparable across runs, we enforce consistency in anchor point configuration. Unlike methods that rely on random uniform sampling, we generate the sampling seed via a hash code derived from the object's unique identifier. This guarantees that each object yields an identical set of $N_A$ anchor points, rendering the predicted physical attributes physically interpretable.

\noindent \textbf{Physical Parameterization and Simulation.}
To facilitate generalization across materials, we bind each 3DGS asset to a spring-mass system and sample physical configurations $\mathbf{p} = (m, k, d, f)$ from continuous distributions: mass $m \in [0.2, 6.0]$, stiffness $k \in [10, 1200.0]$, damping $d \in [0.1, 5.0]$, and friction $f \in [0.0, 1.0]$. This ensures that our dataset covers a continuous distribution of physical behaviors rather than discrete categories. For each $(\mathbf{g}, \mathbf{p})$ pair, we simulate a 30-frame free-fall trajectory under gravity, including ground collision. The resulting motion is rendered into monocular videos $\mathcal{V} = \{I_t\}_{t=0}^T$ at $512 \times 512$ resolution, forming a dataset of $(\mathcal{V}, \mathbf{g}, \mathbf{p})$ triplets.

\subsection{Experimental Setup}
\label{sec:setup}
We construct our dataset from the 496 synthesized objects, splitting them into 450 for training and 46 for testing to evaluate cross-object generalization. Each training object is instantiated with 10 distinct physical samples, while each test object has 2 physical samples reserved for evaluation. The data is rendered as 30-frame monocular videos at $512 \times 512$ resolution from four orthogonal views. Following standard protocols~\cite{springgaus}, the initial 20 frames are used for dynamic reconstruction, and the subsequent 10 frames serve as ground truth for future prediction.

Our method is designed for \textit{single-view generalization}. During both training and testing, the model inputs a single-view 20-frame sequence. In contrast, baseline methods \cite{4dgs, springgaus} follow their original protocols, performing per-scene optimization using 4-view 20-frame inputs for each test object independently. We evaluate visual fidelity using PSNR, SSIM, and LPIPS, and assess 3D geometry accuracy via Chamfer Distance (CD) and Earth Mover's Distance (EMD). For physical disentanglement, we additionally report the Mean Absolute Error (MAE) between predicted and ground-truth physical parameters. All metrics for our method are computed on the held-out single view, while baselines are evaluated on the views used for their optimization.

Regarding implementation, we adopt InternViT-300M~\cite{internvl} as the visual backbone for spatio-temporal feature extraction. The model is trained for 100K iterations with a batch size of 8 on 8 RTX 4090 GPUs. For baseline comparisons, we strictly adhere to their official implementations and optimization protocols.

\begin{figure*}[htbp]
    \centering
    \resizebox{\textwidth}{!}{
        \includegraphics{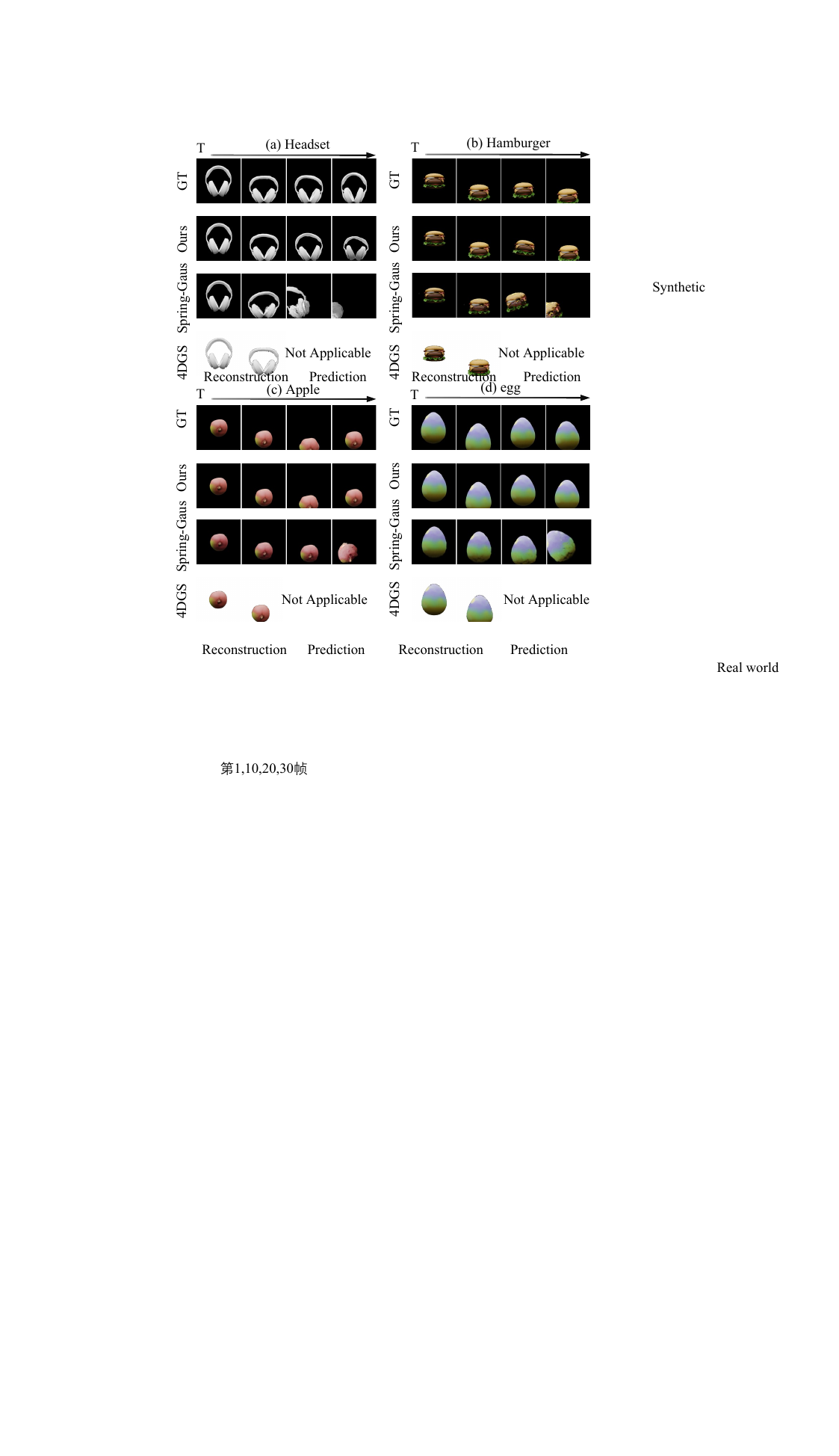}
    }
    \caption{Visualization of different methods. Our method produces more stable dynamics and realistic future states across different objects, while baseline methods either generate unstable deformations or cannot perform future prediction.}
    \label{fig:exp1}
\end{figure*}

\begin{table*}[t]
\centering
\caption{Cross-object generalization on 46 unseen objects. Our method achieves superior future prediction through physics-aware generalization, while maintaining reconstruction quality. Best results in \textbf{bold}.}
\resizebox{\textwidth}{!}{
\begin{tabular}{c|c|c|c|c|c|c|c|c|c|c|c}
\toprule
\multirow{2}{*}{Method} & \multicolumn{5}{c|}{Dynamic Reconstruction} & \multicolumn{5}{c|}{Future Prediction} & Time\\
\cmidrule{2-11}
& PSNR$\uparrow$ & SSIM$\uparrow$ &LPIPS$\downarrow$ & CD$\downarrow$ & EMD$\downarrow$ & PSNR$\uparrow$ & SSIM$\uparrow$ & LPIPS$\downarrow$ & CD$\downarrow$ & EMD$\downarrow$ \\
\midrule
4DGS~\cite{4dgs} & 30.33 & \textbf{0.983} &0.0392& 0.593 & 1.5010 & - & - & - & - & - & $>$1h\\
Spring-Gaus~\cite{springgaus} & 22.26 & 0.874 &0.1517& 0.466 & 0.582 & 13.27 & 0.745 &0.2856& 0.349 & 0.424 & $>$1h\\
\textbf{Ours} & \textbf{33.84} & 0.953 & \textbf{0.0366} & \textbf{0.001} & \textbf{0.007} & \textbf{21.64} & \textbf{0.907} & \textbf{0.0876} & \textbf{0.004} & \textbf{0.017} & \textbf{$<$1s}\\
\bottomrule
\end{tabular}
}
\label{tab:generalization}
\end{table*}

\begin{figure*}[htbp]
    \centering
    \resizebox{\textwidth}{!}{
        \includegraphics{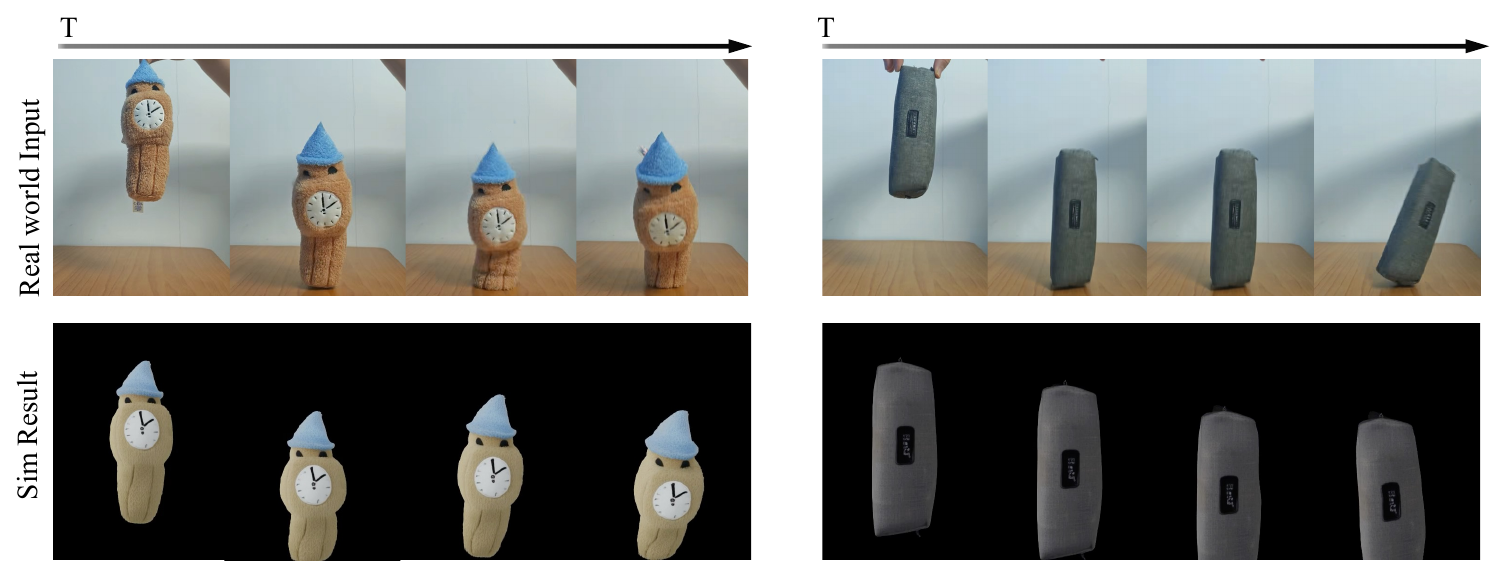}
    }
    \caption{Visualization of real-world non-rigid assets. Two objects are shown dropping and deforming over time T, with real-world input (top) and corresponding simulation results (bottom).}
    \label{fig:real_result}
\end{figure*}

\subsection{Cross-Object Generalization}
We evaluate the model's generalization capability on unseen geometries. As presented in Table~\ref{tab:generalization} and visually corroborated by Figure~\ref{fig:exp1}, our method consistently outperforms baselines on synthesized objects, achieving superior fidelity in both dynamic reconstruction and future prediction. Notably, it attains a PSNR gain of +7.6 dB over Spring-Gaus~\cite{springgaus} in prediction tasks, demonstrating that joint physical-visual learning facilitates robust extrapolation to novel objects. We further validate practical applicability on real-world non-rigid assets in Figure~\ref{fig:real_result}. While quantitative evaluation remains challenging in uncontrolled real-world settings, the qualitative results confirm that our model faithfully captures complex, physics-consistent deformations without per-scene optimization. In contrast to 4DGS~\cite{4dgs}, which lacks explicit physical priors and cannot extrapolate over time, our approach maintains competitive reconstruction quality while enabling predictive simulation. Moreover, it delivers exceptional efficiency, reducing processing time from over 1 hour to under 1 second.

\begin{table*}[t]
\centering
\caption{Physical disentanglement on objects with two distinct attributes sets. Our method preserves geometric fidelity while maximizing trajectory divergence between physical states.}
\resizebox{\linewidth}{!}{
\begin{tabular}{l l c c c c c c c c}
\toprule
Reconstruction & Method & Cylinder-1 & Cylinder-2 & Headset-1 & Headset-2 & Hamburger-1 & Hamburger-2 & Lantern-1 & Lantern-2 \\
\midrule
\multirow{2}{*}{PSNR$\uparrow$} 
& Spring-Gaus~\cite{springgaus} & 20.89 & 20.77 & 23.87 & 21.02 & 26.003 & 21.51 & 20.37 & 19.09 \\
& Ours        & \textbf{36.86} & \textbf{36.35} & \textbf{33.13} & \textbf{31.79} & \textbf{33.63} & \textbf{34.07} & \textbf{33.14} & \textbf{30.86} \\
\midrule
\multirow{2}{*}{SSIM$\uparrow$} 
& Spring-Gaus & 0.8637 & 0.8744 & 0.8993 & 0.8744 & 0.9326 & 0.9104 & 0.8275 & 0.8169 \\
& Ours        & \textbf{0.9833} & \textbf{0.9822} & \textbf{0.9517} & \textbf{0.9435} & \textbf{0.9614} & \textbf{0.9670} & \textbf{0.9473} & \textbf{0.9327} \\
\midrule
\multirow{2}{*}{LPIPS$\downarrow$}
& Spring-Gaus & 0.1880 & 0.1616 & 0.1306 & 0.1424 & 0.0906 & 0.1077 & 0.1840 & 0.1884 \\
& Ours        & \textbf{0.0102} & \textbf{0.0233} & \textbf{0.0297} & \textbf{0.0425} & \textbf{0.0362} & \textbf{0.0339} & \textbf{0.0257} & \textbf{0.0610} \\
\midrule
\multirow{2}{*}{CD$\downarrow$}
& Spring-Gaus & 0.4145 & 0.4845 & 0.4848 & 0.4995 & 0.5324 & 0.5248 & 0.4452 & 0.4609 \\
& Ours        & \textbf{0.0003} & \textbf{0.0005} & \textbf{0.0003} & \textbf{0.0006} & \textbf{0.0003} & \textbf{0.0004} & \textbf{0.0002} & \textbf{0.0027} \\
\midrule
\multirow{2}{*}{EMD$\downarrow$}
& Spring-Gaus & 0.6171 & 0.6749 & 0.6328 & 0.6652 & 0.6299 & 0.6377 & 0.6308 & 0.6560 \\
& Ours        & \textbf{0.0050} & \textbf{0.0054} & \textbf{0.0032} & \textbf{0.0044} & \textbf{0.0034} & \textbf{0.0154} & \textbf{0.0048} & \textbf{0.0128} \\
\bottomrule
\end{tabular}
}
\label{tab:disentanglement}
\end{table*}

\begin{table}[htbp]
\centering
\caption{Comparison of physical attributes errors.}
\resizebox{0.5\linewidth}{!}{
\begin{tabular}{lcccc}
\toprule
Method &  Stiffness & Damp & Mass & Friction  \\
\midrule
Spring-Gaus & 827.67 & 2.546 & 2.276 & \textbf{1.082} \\
Ours & \textbf{297.3} & \textbf{1.151} & \textbf{1.337} & 1.508 \\
\bottomrule
\end{tabular}}
\label{tab:physics_error}
\end{table}
\begin{figure*}[htbp]
    \centering
    \resizebox{0.8\textwidth}{!}{
        \includegraphics{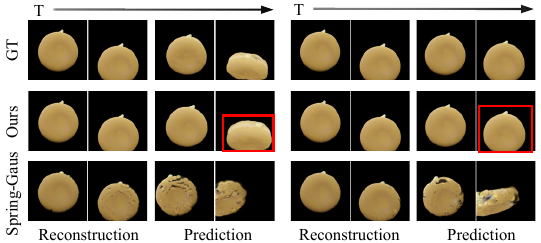}
    }
    \caption{For the same object with different physical attributes, our method produces more accurate deformation predictions compared to Spring-Gaus.}
    \label{fig:exp2}
\end{figure*}
\subsection{Physical Disentanglement}
\label{sec:disentanglement}
To verify the model's ability to disentangle physical attributes from geometric structure, we evaluate 46 objects, each assigned two distinct physical configurations in our dataset. As summarized in Table~\ref{tab:disentanglement}, our method consistently maintains high reconstruction fidelity while accurately distinguishing between these varying physical states. The quantitative results demonstrate that, despite sharing identical underlying geometry, the model successfully captures and separates the distinct physical properties assigned to each instance, confirming its strong capability in geometry-attribute decoupling.

To quantitatively validate this disentanglement, we further evaluate the accuracy of physical property recovery by computing parameter estimation errors. As detailed in Table~\ref{tab:physics_error}, our approach not only outperforms Spring-Gaus in reconstruction quality and geometric fidelity but also achieves substantially lower errors in physical parameter estimation. This marked reduction in error confirms that our model successfully decouples visual appearance from underlying physics. Consequently, identical geometries assigned distinct material properties yield accurately differentiated physical predictions, while the visual reconstruction remains consistently high-fidelity.

Figure~\ref{fig:exp2} visualizes this capability, illustrating divergent future trajectories when the same object is simulated with different predicted $\mathbf{p}$. These results validate that our self-supervised training strategy effectively learns to infer physically meaningful attributes from monocular video alone, without requiring explicit physics supervision.

\begin{figure*}[htbp]
    \centering
    \resizebox{\textwidth}{!}{%
        \includegraphics{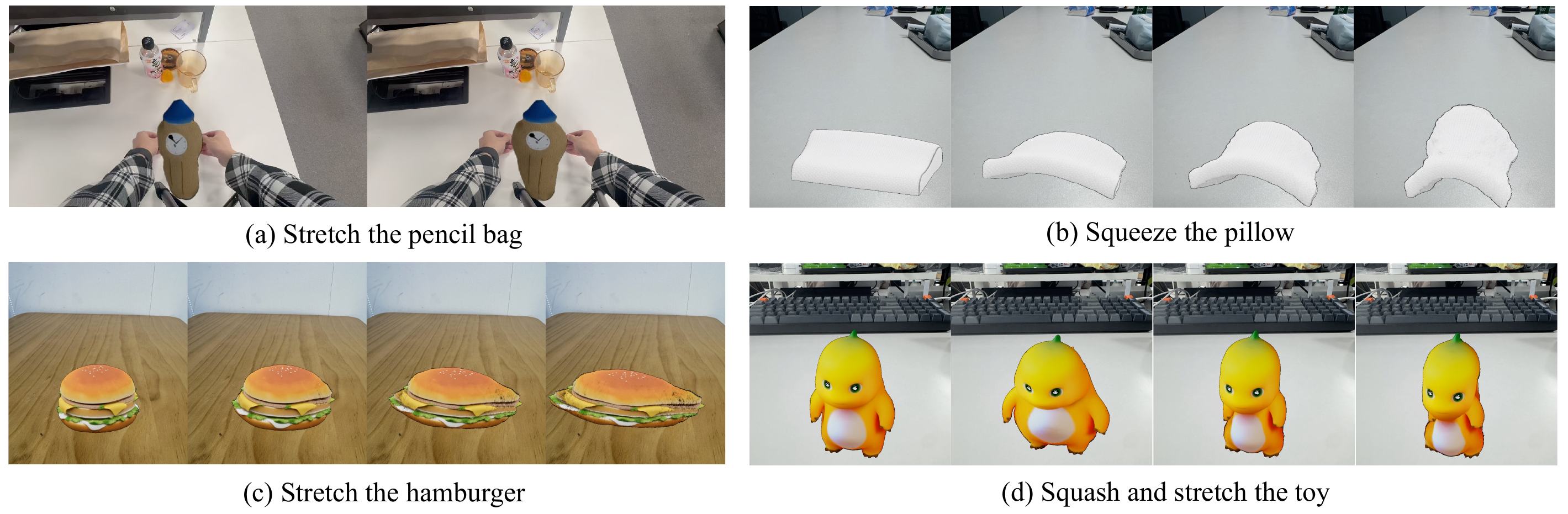}
    }
    \caption{Our method can be utilized in non-rigid manipulation. Four manipulation scenarios are demonstrated: (a) stretching the pencil bag, (b) squeezing the pillow, (c) stretching the hamburger, and (d) squashing and stretching the toy, showcasing the versatility of our approach across diverse deformable objects.}
    \label{fig:manipu_demo}
\end{figure*}

\subsection{Application in Robot Non-rigid Object Manipulation}
This methodology demonstrates the applicability of video-based non-rigid object reconstruction to robotic deformable object manipulation. As illustrated in Figure~\ref{fig:manipu_demo}, we first capture a video sequence of a freely falling object. The target object is segmented using the Segment Anything Model (SAM)~\cite{sam}, and the segmented video is subsequently processed by ReconPhys to reconstruct both a 3DGS representation and a spring-mass system. Within a virtual simulation environment, configured following the framework of PhysTwin~\cite{phystwin}, controller points can be interactively manipulated via keyboard inputs to simulate robotic gripper or anthropomorphic hand interactions with the deformable object. This pipeline provides an efficient solution for rapidly acquiring high-fidelity deformable digital assets tailored for photorealistic physical simulation, thereby establishing a practical pathway toward scalable simulation frameworks for robotic manipulation of non-rigid objects.

\section{Conclusion}
\label{sec:conclusion}

We introduce a self-supervised framework that jointly infers 3D Gaussian Splatting representations and physical attributes of deformable objects from monocular free-fall videos. By establishing end-to-end differentiable pathways between visual reconstruction loss and physical simulation, our dual-branch architecture eliminates the need for ground-truth physics labels while achieving superior cross-object generalization and physical disentanglement. Experiments demonstrate that modeling dynamics through gradient flow to physical attributes acts as an effective regularizer, yielding 8.37dB higher PSNR in future prediction than appearance-only baselines and clear trajectory divergence for objects with identical geometry but different material properties. This work bridges visual perception and physical understanding for dynamic object modeling.

\clearpage
\setcitestyle{numbers}
\bibliographystyle{plainnat}
\bibliography{main}

\end{document}